\documentclass[runningheads]{llncs}

\usepackage[T1]{fontenc}
\usepackage{amsmath}
\usepackage{booktabs}
\usepackage{amsfonts}
\usepackage{enumitem}
\usepackage[style=ieee,sorting=none]{biblatex}
\usepackage[colorlinks=true, citecolor=blue, linkcolor=black, urlcolor=blue]{hyperref}
\usepackage{xcolor} 

\usepackage{etoolbox}
\makeatletter
\pretocmd{\ref}{\textcolor{blue}}{}{}
\makeatother

\usepackage{graphicx}

\begin{document}

\title{SignBart - New approach with the skeleton sequence for Isolated Sign language Recognition}
\titlerunning{SignBart}

\author{ 
Tinh Nguyen \inst{1}\orcidID{0009-0003-1474-0691}\and
Minh Khue Phan Tran\inst{1}\orcidID{0009-0007-7711-9586}}

\authorrunning{Tinh Nguyen \and Minh Khue Phan Tran}

\renewbibmacro*{volume+number+eid}{%
  \printfield{volume}%
  \setunit*{\addcomma\space}%
}

\institute{Ho Chi Minh Open University, VietNam 
\email{ou@ou.edu.vn}\\
\url{https://ou.edu.vn/}}

\setlength{\parindent}{2em}
\setlength{\topsep}{0pt}
\setlength{\partopsep}{0pt}

\maketitle             
\begin{abstract}
Sign language recognition is crucial for individuals with hearing impairments to break communication barriers. However, previous approaches have had to choose between efficiency and accuracy. Such as RNNs, LSTMs, and GCNs, had problems with vanishing gradients and high computational costs. Despite improving performance, transformer-based methods were not commonly used. This study presents a new novel SLR approach that overcomes the challenge of independently extracting meaningful information from the x and y coordinates of skeleton sequences, which traditional models often treat as inseparable. By utilizing an encoder-decoder of BART architecture, the model independently encodes the x and y coordinates, while Cross-Attention ensures their interrelation is maintained. With only 749,888 parameters, the model achieves 96.04\% accuracy on the LSA-64 dataset, significantly outperforming previous models with over one million parameters. The model also demonstrates excellent performance and generalization across WLASL and ASL-Citizen datasets. Ablation studies underscore the importance of coordinate projection, normalization, and using multiple skeleton components for boosting model efficacy. This study offers a reliable and effective approach for sign language recognition, with strong potential for enhancing accessibility tools for the deaf and hard of hearing.

\end{abstract}

\section{Introduction}\label{sec:introduction}
Developed to help Deaf people communicate, sign language is a graphically structured language system with special syntactic and morphological characteristics\cite{Othman2024}. It requires complex coordination between several visual cues and combines both manual (hand movements, body gestures) and non-manual (facial expressions, head movements) components\cite{mukushev2020evaluation}. Sign language presents substantial computational modeling issues due to its multimodal character, especially in Deep Learning applications\cite{chaggle_slr}. Sign language is still a vital tool for improving accessibility and inclusion in human communication, as an estimated 5\% of the
world’s population suffers from hearing loss, sign language remains an essential tool for enhancing accessibility and inclusivity in human communication \cite{who2021deafness}.

To enhance accessibility while supporting social inclusion, scientific research has become very interested in Sign Language Recognition - SLR\cite{shiri2023comprehensive}. Two main tasks are involved in SLR:
\begin{itemize}[label=\textbullet, left=1cm, topsep=0pt, partopsep=0pt]
  \item Isolated Sign Language Recognition - ISLR, where each video corresponds to a word in sign language \cite{grobel1997isolated}.
  \item Continuous Sign Language Recognition - CSLR, where each video consists of a sequence of sign language forming a sentence \cite{aloysius2020understanding}.
\end{itemize}

SLR developed in a variety of ways, from LSTMs \cite{cnn-lstm, lstm-cslr} and RNNs\cite{mediapipe-rnn, rnn-transducer} to Transformers\cite{signbert, spoter} and GCNs\cite{sl-gcn, spatial-gcn}. However, a persistent problem for these approaches is computational complexity and the insufficient use of x-y coordinate relationships in skeleton data\cite{skeletons__Delaunay_tri, skelemotion}. The focus of GCN-based models is on keypoint movements, but as graph complexity increases, they are prone to overfitting and struggle to capture long-range interactions\cite{gcn_app, gcn_survey}. Transformers \cite{signbert, spoter}, on the other hand, use attention to improve contextual learning, but they are not commonly used. 

The goal of this study is to get around these problems by suggesting a model based on the BART architecture, which is both accurate and has a low computation cost. BART \cite{bart}, based on Facebook's Transformer design, encodes sequential data bidirectionally, a significant advantage over unidirectional models like LSTM or GRU. The model leverages Self-Attention and Self-Causal-Attention mechanisms to capture bidirectional context and improve data generalization.  

\vspace{1px}
Main contributions of the study:
\begin{itemize}[label=\textbullet, left=1cm, topsep=0pt, partopsep=0pt]
    \item Proposing a new approach using skeleton data in ISLR.
    \item Contributes a lightweight model that balances complexity and accuracy on ISLR datasets.
    \item Analyzing the effectiveness of the proposed approach in the study.
\end{itemize}

\section{Related Work}\label{sec:related_work}
ISLR has employed a variety of methods, including the analysis of images and videos with handcrafted features (e.g., HOG, SIFT)\cite{shit-BSL, SL-HOG}, to identify sign language gestures. Nevertheless, the adaptability of these methods is restricted by the fact that they are unable to generalize effectively across various contexts, as the features are not acquired from the data\cite{disadvantages_of_shift_hog}. On the other hand, Deep Learning (DL) allows models to independently extract features from data, facilitating the recognition of intricate patterns, such as sign language gestures, without human involvement\cite{survey_slr}. The result has greatly contributed to the progression and accuracy of ISLR research.
Video-based\cite{al2021deep} and pose-based\cite{al2021deep} are two approaches into which deep learning has developed a variety of methods for sign language recognition.

\textbf{Video-based Method:} Initial deep learning methodologies employed Convolutional Neural Networks (CNNs) to extract features from RGB or RGB+D data. Initially, 2D CNNs extracted spatial characteristics from images\cite{cnn2d-in-AR}, whereas 3D CNNs enhanced this capability by including temporal filters, facilitating the acquisition of dynamic features\cite{3dcnn_ar}. This development has enhanced action recognition in SLR, \cite{aslcitizen} retrained I3D models on datasets on ASL Citizen to enhance accuracy. \cite{vitrans_islr} trained the VideoMAE, SVT, MakeFast, and BEVT models using the WLASL2000 dataset. In the study \cite{gsl_iso}, the authors utilized SubUNets, GoogLeNet+Tconvs, 3D-ResNet, and I3D architectures on the GSL-iso dataset. The results of the approaches are accurate. However, the model's complexity and computational cost are high.

\textbf{Pose-based methods:} Pose sequences extracted from video serve as an efficient data structure for ISLR, capturing movement and reducing computation costs. The study in \cite{st-gcn} first used posture sequences for action recognition, using Graph Convolutional Networks (GCN) to improve the model's learning of skeleton movements over time and execute classification. More research studies have shown the effectiveness of GCN in action recognition. GCN has also been investigated and used for Isolated SLR. \cite{sl-gcn} proposed SL-GCN designs that integrate pose sequence data with skeleton graph representations for ISLR. Transformer-based has used \cite{spoter} introduced the SPOTER architecture, wherein the Encoder processes the posture sequence, and the Decoder incorporates a parameter known as the class query, removing Self-Causal Attention. The study \cite{signbert} introduced the SignBERT architecture, based on BERT, for solving the problems of ISLR and CSLR. The results from this research show that utilizing posture sequences enables the models to get higher accuracy and reduced complexity in comparison with RGB+D video approaches.

\section{Approach}\label{sec:approach}
\subsection{Keypoints approach}\label{sec:keypoints_approach}
Earlier methods used RNNs\cite{rnn-transducer, mediapipe-rnn}, but vanishing gradients limited their effectiveness. LSTMs\cite{lstm-cslr, cnn-lstm} solved this challenge but had high computation costs because of using multiple gates. ST-GCN\cite{st-gcn} circumvented these constraints by using graph-based architectures to represent skeleton movements. Even though GCNs work well, they get expensive to run on computers as graph complexity increases, and they often overfit because they are sensitive to motion graph parameters\cite{comprehensive_gcn}. SL-GCN\cite{sl-gcn} enhanced this by using multi-stream input but had a high computational cost. The challenge for GCNs is to limit the interchange of information across vast distances\cite{comprehensive_gcn}. Transformer\cite{trans_based} were applied, and  SignBERT\cite{signbert} achieved good results despite their computational demands. SPOTER \cite{spoter} reduced complexity and achieved 100\% accuracy on LSA-64 by replacing Self-Causal-Attention with class-query Cross-Attention. However, class-query struggled on large datasets due to inconsistent gradient updates, which led to convergence issues. Although transformer-based and GCN models have succeeded significantly, their limitations imply a trade-off between computer efficiency and accuracy in SLR research.

In the past, models thought that the x and y axes were strongly connected based on the theory of coordinate relationships in skeleton data\cite{chaggle_slr}. In a 2D space, each skeleton keypoint is defined by both the x (horizontal) and y (vertical) coordinates \cite{coordinate_theory}. Consequently, these models handle every data point as an inseparable pair of (x, y) values, where a point can only be specified in the presence of both x and y. Although each axis has unique properties and functions in expressing bodily motion, this results in models that concurrently encode the two axes as a single object. Nevertheless, especially if working with complex skeleton data, this approach can make it difficult to distinguish and separately extract information from each axis, making it more difficult to comprehend the unique characteristics of each.

In this study, a new approach is proposed based on the theory that x and y coordinates lie on two distinct axes but share a special relationship\cite{coordinate_theory}. The study applies the Encoder-Decoder structure of the BART \cite{bart}. With separate encoding, the encoder will encode the x coordinates while the decoder will encode the y coordinates. With distinct encoding, the encoder will encode the x coordinates while the decoder will encode the y coordinates. This way, the model is able to comprehend each pair of values independently. X and Y, due to their theoretical interdependence, might lose information if encoded independently. To deal with this, the decoder uses and updates the y encoding with data from the x coordinates via Cross-Attention with the encoded x values from the encoder while encoding y. This approach addresses the two main problems with the theory of coordinate connections.

\subsection{Model Architecture}\label{sec:model_arch}
\begin{figure}[h]
    \centering
    \includegraphics[width=300px]{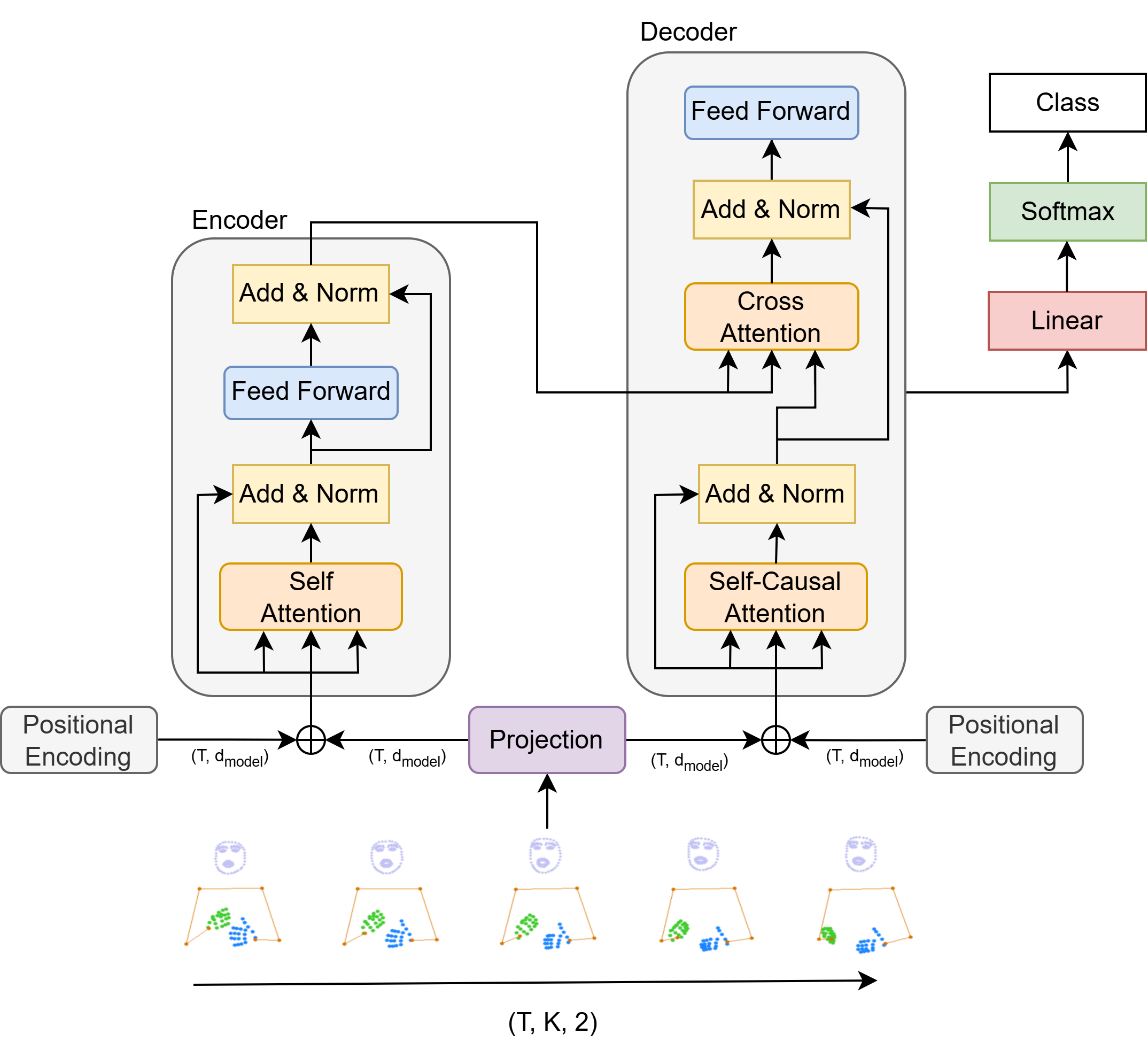}
    \caption{Model architecture. With the input skeleton data, the Encoder encodes the x
 coordinate, and the Decoder encodes the y coordinate and query information from the
 encoded x coordinate. Before encoding, both the x and y coordinates will go through
 a mapping process via Projection.}
    \label{fig:model_arch}
\end{figure}

\subsubsection{Overview}\label{sec:overview}

Figure \ref{fig:model_arch} illustrates that SignBart comprises Encoder and Decoder block. The input skeleton sequence \( I \in \mathbb{R}^{T \times K \times 2} \) which \( T \) frames, \( K \) keypoints, and two coordinates (x, y). The Encoder encode the x coordinates (\( I_x \)) while the Decoder encode the y coordinates (\( I_y \)) along with the encoded \( I_x \). Before entering the network, both \( I_x \) and \( I_y \) are projected to \( d_{\text{model}} \) via separate Linear layers and enriched with Positional Encoding. Finally, a Linear layer with softmax produces the predictions.
\subsubsection{Projection}\label{sec:projection}

Before being encoded by Encoder and Decoder, \(I_x\) and \(I_y\) are mapped to \(d_{\text{model}}\), similar to token embedding in NLP. Given an input \(I \in \mathbb{R}^{T \times K \times 2}\) (with \(T\) frames, \(K\) keypoints, and 2 representing x and y coordinates), and

\[
x_{\text{coord}} = I_{(:, :, 0)}, \quad y_{\text{coord}} = I_{(:, :, 1)}
\]

Each coordinate is then linearly mapped:

\[
x_{\text{emb}} = x_{\text{coord}} \cdot W_x + b_x, \quad y_{\text{emb}} = y_{\text{coord}} \cdot W_y + b_y,
\]

where \(W_x, W_y \in \mathbb{R}^{K \times d_{\text{model}}}\) and \(b_x, b_y \in \mathbb{R}^{d_{\text{model}}}\). 
\\ \\
The resulting embeddings are:
$
x_{\text{emb}},\, y_{\text{emb}} \in \mathbb{R}^{T \times K \times d_{\text{model}}}.
$

\subsubsection{Self Attention}\label{sec:self_attention}
After mapping the coordinates to \( d_{\text{model}} \) via Projection and adding positional information through Positional Encoding, the x-coordinate sequences are encoded by the Encoder’s Self-Attention , allowing sequences to interact and capture the bidirectional information of them. An attention mask marks valid sequences and padding.
    
The input, represented as \( H \in \mathbb{R}^{T \times d_{\text{model}}} \), is linearly transformed into queries (\( Q \)), keys (\( K \)), and values (\( V \)) using weight matrices \( W_Q, W_K, W_V \in \mathbb{R}^{d_{\text{model}} \times d_{\text{model}}} \). Each head computes attention as:

\[
O_i = \text{softmax}\left( \frac{Q_i K_i^T}{\sqrt{d_{\text{head}}}} + M \right) V_i
\]

Where \( M \) is the attention mask, ensuring padding sequences do not affect the computation. The outputs from all heads are concatenated and transformed via \( W_O \), followed by dropout to prevent overfitting. This mechanism enhances the model’s ability to capture relational patterns and efficiently process skeleton sequences.

\subsubsection{Self-Causal-Attention}
\label{sec:self_causal_attention}

For the y-coordinate embedding of the skeleton sequence, the computational procedure for generating the attention map is performed similarly to the x-coordinate embedding in Self-Attention in section \ref{sec:self_attention}. The main difference is that a causal mask allows each sequence to interact and capture only itself and the previous sequences. This enhances the efficiency of querying in Cross-Attention.
The causal mask \( M \in \mathbb{R}^{T \times T} \) is defined as

\[
M_{(i,j)} =
\begin{cases} 
1, & \text{if } i \geq j \\ 
0, & \text{if } i < j 
\end{cases}, \text{ for } i, j \in \{0, 1, \dots, T\}
\]

\subsubsection{Cross Attention}\label{sec:cross_attention}
After generating the attention map via Self-Causal-Attention in section \ref{sec:self_causal_attention}, Cross-Attention links the x-coordinate embeddings from the Encoder with the Self-Causal-Attention map. This integration captures positional dependencies between x and y coordinates for each skeleton frame.

The process follows Self-Attention in section \ref{sec:self_attention}, but differs in how \( Q \), \( K \), and \( V \) are derived:

\[
Q = A_y \cdot W_Q, \quad K, V \text{ from } A_x
\]

Here, \( Q \) comes from the Self-Causal-Attention map (\( A_y \)), while \( K \) and \( V \) originate from the Encoder's attention map (\( A_x \)). This formulation enriches the representation by aligning x and y dependencies.

\subsection{Extract Keypoints}\label{extract_keypoints}

Using Google’s Mediapipe, keypoints are extracted from sign language video frames, including the body, left hand, and right hand. Mediapipe extracts 33 body keypoints, and 21 for each hand, but only 6 body keypoints are used. Each keypoint consists of two 2D coordinates (x, y).

The extracted for each video has the shape \((T, 75, 2)\), where: \(T\) is the number of frames, 75 is the total keypoints per frame (\(6\) body + \(21\) left hand + \(21\) right hand), 2 represents the x and y coordinates.

Missing keypoints are assigned coordinates of 0. To normalize the coordinate keypoints to range \([0, 1]\) according to the formula:

\[
x = \frac{x}{W}, \quad y = \frac{y}{H}
\]

Which \(W\) and \(H\) are the frame's width and height.

\subsection{Normalization}\label{normalization}
The keypoints in a skeleton sequence are influenced by the signer's position in the video. Without normalization, variations caused by factors like camera distance and tilt would lead to vastly different coordinates for the same sign, making it harder for the model to generalize. This would increase training time and hinder the model's ability to learn relevant patterns.

To address this, three main parts of the body are considered: the body, the left hand, and the right hand. A bounding box is created for each part by calculating the top-left and bottom-right corners, with a 5\% margin added to ensure the keypoints are fully encompassed. Normalization is then applied using the following formula:

\[
x = \frac{x - x_{\min}}{x_{\max} - x_{\min}}, \quad y = \frac{y - y_{\min}}{y_{\max} - y_{\min}}
\]

where:  
\begin{itemize}[label=\textbullet, left=1cm]
 \item \((x, y)\) are the raw keypoint coordinates.  
 \item \((x_{\min}, y_{\min})\) are the coordinates of the top-left corner of the bounding box.  
 \item \((x_{\max}, y_{\max})\) are the coordinates of the bottom-right corner of the bounding box.  
\end{itemize}
So, each part is normalized independently based on its local width and height, making the data independent of the frame size. Enhancing generalization reduces training time, and improves model accuracy by making the learning process more efficient and robust.

\section{Experiments}

This section presents the implementation and evaluation of the study approach, including the model architecture, training setup, and datasets used. Importantly, the study compares SignBart's results with those of state-of-the-art models and investigates the influence of various factors through ablation studies.

\subsection{Implementation details} 
The model consists of 2 Encoder and 2 Decoder blocks, each with 16 attention heads, enabling efficient representation learning in skeleton sequences. The dimensions \( d_{\text{model}} \) and \( ff_{\text{dim}} \) are adjusted per dataset to optimize efficiency and prevent overfitting. Training employs the AdamW optimizer with a weight decay of \( 1 \times 10^{-2} \). The learning rate starts at \( 2 \times 10^{-4} \) and follows a cosine annealing schedule with warmup. A batch size of 128 balances memory use and convergence.

\subsection{Datasets}
The model is trained and evaluated on the datasets LSA64 \cite{lsa-64}, ASL-Citizen \cite{aslcitizen}, and WLASL\cite{wlasl}. These datasets provide various challenges and cover a wide range of sign language gestures, enhancing the model's ability to generalize across different types of sign language data. Basic information about the datasets used in the study is summarized in Table \ref{tab:sign_language_datasets}

\begin{table}
\caption{An overview of Datasets was used in study}
\centering
\begin{tabular}{|l|l|l|l|l|}
\hline
\textbf{Dataset}    & \textbf{Number of Videos} & \textbf{Number of Gloss} & \textbf{Signers} & \textbf{Language}     \\ \hline
\textbf{WLASL}      & 21083                    & 2000                     & 119               & American             \\ \hline
\textbf{LSA-64}     & 3200                     & 64                       & 10                & Argentinian          \\ \hline
\textbf{ASL-Citizen}& 84000                    & 2731                     & 52                & American             \\ \hline
\end{tabular}
\label{tab:sign_language_datasets}
\end{table}

\subsubsection{WLASL\cite{wlasl}}
(Word-Level American Sign Language) is a sign language dataset for American Sign Language that was developed to help studies for sign language recognition. Contains 21,083 videos of 2000 words taken from various internet sources. WLASL has been split into four subsets based on the number of words and the level of complexity, giving a comprehensive test of model performance in different contexts.

\subsubsection{LSA-64 \cite{lsa-64}}
The LSA64 dataset,which focuses on 64 commonly used words in Argentinian Sign Language (LSA), comprises 3200 videos produced by 10 non-expert signers. Both verbs and nouns from frequently used entries in the LSA dictionary were the source of the chosen phrases.

\subsubsection{ASL-Citizen \cite{aslcitizen}}
With 2,731 words in over 84K videos, ASL-Citizen is the first dataset created by American Sign Language Communication. Created by 52 deaf or hard-of-hearing people, the videos were created using a community-driven sign language platform. Similar to WLASL, ASL-Citizen is split into several variants with 100, 200, 400, 1000, and 2731 words to assess the model's effectiveness in different contexts.

\subsection{Comparison with State-of-the-art Methods}

\textbf{WLASL\cite{wlasl}:} The performance of the approach in study on the subsets of WLASL is shown in Table \ref{tab:eval_wlasl}. The NLA-SLR method \ is considered state-of-the-art for WLASL. As shown in Table \ref{tab:compared_nlaslr}, NLA-SLR\cite{NLA-SLR} achieves high accuracy on WLASL-100 and WLASL-300. However, to achieve these high accuracies, the model must process two types of input data: RGB and skeleton joint sequences, which complicates the model architecture. This complexity becomes a challenge for WLASL-1000 and WLASL-2000, as larger datasets require the model to generalize well to differentiate many classes. In contrast, the approach in this study maintains stable accuracy across all four WLASL subsets and demonstrates superior generalization capability, achieving a 5.73\% increase in accuracy on WLASL-300 and a 9.69\% increase on WLASL-2000. 

\begin{table}
\caption{Comparison with Other Models on WLASL\cite{wlasl} Subsets with top-1 accuracy}
\label{tab:eval_wlasl}
\centering
\begin{tabular}{|l|c|c|c|c|}
\hline
\textbf{Model}       & \textbf{WLASL-100} & \textbf{WLASL-300} & \textbf{WLASL-1000} & \textbf{WLASL-2000} \\ \hline
I3D\cite{wlasl}                  & 65.89\%                   & 56.14\%                    & 47.33\%                     & 32.48\%                      \\ \hline
Fusion-3\cite{Fusion-3}             & 75.67\%                   & 68.30\%                    & 56.68\%                     & 38.84\%                      \\ \hline
BEST\cite{best}                 & 81.63\%                   & 76.12\%                    & -                           & 52.12\%                      \\ \hline
SignBERT\cite{signbert}             & 82.56\%                   & 74.40\%                    & -                           & 52.08\%                      \\ \hline
NLA-SLR \cite{NLA-SLR}
& \textbf{93.08\%}                   & \textbf{87.33\%}                    & 75.72\%                     & 58.31\%                      \\ \hline
SPOTER\cite{spoter}               & 63.18\%                   & 43.78\%                    & -                           & -                            \\ \hline
SignBart            & 78.00\%                   & 78.50\%                    & \textbf{81.45\%}                     & \textbf{68.00\%}                      \\ \hline
\end{tabular}
\end{table}
\begin{table}
\caption{Detailed Comparison with SOTA Model (NLA-SLR\cite{NLA-SLR}) with top-1 accuracy}
\label{tab:compared_nlaslr}
\centering
\begin{tabular}{|l|c|c|c|c|}
\hline
\textbf{Subset} & \textbf{NLA-SLR\cite{NLA-SLR}} & \textbf{Parameters} & \textbf{SignBart} & \textbf{Parameters} \\ \hline
WLASL-100       & \textbf{93.08\% }                & 84,511,404          & 78.00\%                  & 755,556            \\ \hline
WLASL-300       & \textbf{87.33\%}                 & 89,429,404          & 78.50\%                  & 2,873,132          \\ \hline
WLASL-1000      & 75.72\%                 & 106,642,404         & \textbf{81.45\%}                  & 3,578,344          \\ \hline
WLASL-2000      & 58.31\%                 & 131,232,404         & \textbf{68.00\% }                 & 3,835,344          \\ \hline
\end{tabular}
\end{table}

\newpage
\textbf{LSA-64\cite{lsa-64}:} As shown in Table \ref{tab:eval_lsa64}, the methods \cite{spoter}, \cite{HWGATE}, \cite{sl-gcn}, and \cite{3dgcn} have achieved very high accuracy, all above 90\%. Notably, SPOTER\cite{spoter} achieved 100\% accuracy. However, previous models had more than one million parameters. The approach in this study improves the model's complexity, with only 749,888 parameters, much lower than previous models, but it still demonstrates superior effectiveness, achieving 96.04\% accuracy, which is higher than ST-GCN\cite{sl-gcn} and 3DGCN\cite{3dgcn}.

\begin{table}
\caption{Comparison with the state-of-the-art methods in Top-1 accuracy on the LSA-64\cite{lstm-cslr}. 3DGCN\cite{3dgcn} doesn't publish code, so can't get its parameters.}
\label{tab:eval_lsa64}

\centering
\begin{tabular}{|l|l|l|}
\hline
\textbf{Model}       & \textbf{Validation (Acc)} & \textbf{Parameters} \\ \hline
Spoter\cite{spoter}               & \textbf{100\%}                     & 5,918,848                     \\ \hline
HWGATE\cite{HWGATE}               & 98.59\%                   & 10,758,354                    \\ \hline
ST-GCN\cite{HWGATE}               & 92.81\%                   & 3,604,180                     \\ \hline
SL-GCN\cite{sl-gcn}               & 98.13\%                   & 4,872,306                     \\ \hline
3DGCN\cite{3dgcn}               & 94.84\%                   & -                             \\ \hline
SignBart            & \textbf{96.04\%} & \textbf{749,888} \\ \hline
\end{tabular}
\end{table}

\textbf{ASL-Citizen\cite{aslcitizen}:} is a dataset published in 2023, and to date, there has not been a benchmark comparison on ASL-Citizen. Apart from the pretraining model from the ASL-Citizen paper\cite{aslcitizen}, no studies on ISLR have been conducted on ASL-Citizen. ST-GCN and I3D are the models trained in the original paper. Both models achieved success in action recognition, but when applied to ASL-Citizen, where the number of glosses reaches 2731, as shown in Table \ref{tab:eval_asl-2731}, these two models still do not demonstrate high data generalization, despite their large number of parameters. SignBart, on the other hand, has a lot fewer parameters but is better at generalizing data, as shown by the fact that it is more accurate than the first two models.

\begin{table}
\caption{Results of SignBart on ASL-Citizen-(class).}
\label{tab:eval_asl_dataset}
\centering
\begin{tabular}{|l|l|l|}
\hline
\textbf{ASL-Citizen (class)} & \textbf{Validation} & \textbf{Parameters} \\ \hline
ASL-Citizen-100             & 80.32\%                   & 754,532                      \\ \hline
ASL-Citizen-200             & 81.49\%                   & 2,845,384                    \\ \hline
ASL-Citizen-400             & 78.96\%                   & 3,424,144                    \\ \hline
ASL-Citizen-1000            & 81.45\%                   & 3,578,344                    \\ \hline
ASL-Citizen-2731            & 75.22\%                   & 4,548,523                   \\ \hline
\end{tabular}
\vspace{8px}
\caption{Comparison with two pre-trained models in original paper\cite{aslcitizen}}
\label{tab:eval_asl-2731}
\centering
\begin{tabular}{|l|c|c|c|}
\hline
\textbf{Model}       & \textbf{Rec@1} & \textbf{Rec@5} & \textbf{Parameters} \\ \hline
I3D                  & 63.10\%        & 86.09\%    & 15,086,539                    \\ \hline
ST-GCN               & 59.52\%        & 82.68\%   & 3,788,165                     \\ \hline
SignBart            & \textbf{75.22\%}        & -              & \textbf{4,548,523}   \\ \hline
\end{tabular}
\end{table}

\subsection{Ablation Study}

To evaluate the role of different components in the model, ablation experiments were conducted to identify the clear impact of each factor on model performance. The results show that carefully choosing and adjusting things like data preprocessing and deep learning mechanisms not only makes the model more accurate but also makes it better at applying what it has learned to new situations.

\subsubsection{Projection}

Table~\ref{tab:projection_effect} shows the impact of mapping coordinate systems before the Encoder and Decoder. Without mapping, attention vectors have limited meaning, restricting the model's ability to utilize sequence information. Mapping expands these vectors, enhancing accuracy, with the model achieving 96.04\% in this study.

\begin{table}
\caption{Effect of Projection on validation split the LSA-64\cite{lstm-cslr}}
\label{tab:projection_effect}
\centering
\begin{tabular}{|l|c|}
\hline
\textbf{Projection} & \textbf{Top-1-accuracy} \\ \hline
No projection       & 62.08\%               \\ \hline
With projection     & \textbf{96.04\%}               \\ \hline
\end{tabular}
\end{table}

\subsubsection{Normalization}

To evaluate the effectiveness of normalization, the model was trained with four different versions of normalization on LSA-64. With an accuracy increase of 13.54\% when applying normalization, as shown in Table \ref{tab:normalization_effectiveness}, this highlights the importance of normalization in contributing to the success of ISLR.

\begin{table}
\caption{Impact of Normalization Effect on validation split the LSA-64\cite{lsa-64}. Note: one bounding box (body + left hand + right hand), two bounding boxes (body, left hand + right hand), and three bounding boxes (body, left hand, right hand) }
\label{tab:normalization_effectiveness}
\centering
\begin{tabular}{|l|c|}
\hline
\textbf{Normalization} & \textbf{Top-1-Accuracy} \\ \hline
No                    & 82.50\%               \\ \hline
One bounding box      & 90.52\%               \\ \hline
Two bounding boxes     & 90.41\%               \\ \hline
Three bounding boxes   & \textbf{96.04\%}               \\ \hline
\end{tabular}
\end{table}

\subsubsection{Skeleton Components}

Table \ref{tab:skeletal_components} shows the impact of each skeleton component. Using individual components leads to suboptimal results: the body achieves 86.97\%, the left hand 23.02\%, the right hand 70.20\%, and both hands combined 91.35\%. The best performance (96.04\%) comes from combining all three parts. The study also highlights that signers rely more on their right hand. This is evident as right-hand keypoints alone achieve 70.20\% accuracy, while the left hand only reaches 23.02\%, likely due to right-hand dominance in sign language.

\begin{table}
\caption{Effect of Skeleton Components on LSA-64}
\label{tab:skeletal_components}
\centering
\begin{tabular}{|l|l|l|c|}
\hline
\textbf{Body} & \textbf{Lefthand} & \textbf{Righthand} & \textbf{Test Accuracy} \\ \hline
X             &                   &                    & 86.97\%               \\ \hline
              & X                 &                    & 23.02\%               \\ \hline
              &                   & X                  & 70.20\%               \\ \hline
X             & X                 &                    & 91.35\%               \\ \hline
X             & X                 & X                  & \textbf{96.04\%}               \\ \hline
\end{tabular}
\end{table}

\newpage
\section{Conclusion}

The study introduces an approach for the ISLR model that leverages spatial correlations in skeleton data. Unlike previous approaches that treated x and y coordinates as inseparable pairs, the model encodes them independently while maintaining their interdependence via Cross-Attention. This Encoder-Decoder architecture achieves state-of-the-art accuracy with fewer parameters, outperforming prior models. Evaluated on LSA-64, WLASL, and ASL-Citizen, SignBart achieves 96.04\% accuracy on LSA-64 and shows superior generalization on subsets of WLASL and ASL-Citizen, where previous models struggled with complexity and overfitting. Ablation studies highlight the importance of normalization, multi-part skeleton input (body, left hand, right hand), and coordinate mapping. SignBart represents a major advancement in ISLR, balancing accuracy and efficiency. Its results pave the way for more readable, scalable skeleton-based models, facilitating real-world applications like improved accessibility tools for the deaf.

However, the studies still have limitations: the encoding of x and y coordinate values before queries may not fully reflect the actual spatial relationship and dynamics of the gesture, losing important information. Moreover, three Attention machines could increase the computational cost, especially with datasets containing a lot of keypoints, impacting the computational cost when applied on mobile devices. Finally, this method was evaluated only on the ISLR and has not been evaluated on the CSLR, where the model needs to accurately recognize each gloss in a video for creating a sentence. Consequently, to improve the generalizability and practical utility, the approach requires additional investigation on mobile devices and CSLR.

\printbibliography{}

\end{document}